\documentclass[11pt,a4paper]{article}
\usepackage[hyperref]{acl2021}
\usepackage{times}
\usepackage{latexsym}

\usepackage{microtype}
\usepackage{amsmath}
\usepackage{graphicx}
\usepackage{color}
\usepackage{multirow}
\usepackage{float}
\usepackage{diagbox}
\usepackage{array}
\usepackage{booktabs}
\usepackage{hhline}
\usepackage{makecell}
\usepackage{caption, subcaption}
\usepackage{tabularx}
\usepackage{amssymb}
\usepackage[linesnumbered,ruled,vlined]{algorithm2e}

\aclfinalcopy 

\SetKwInput{KwInput}{Input}                
\SetKwInput{KwOutput}{Output}              

\title{Knowledgeable or Educated Guess? Revisiting Language Models as Knowledge Bases}

\author{Boxi Cao${}^{1,3}$, Hongyu Lin${}^{1}$\thanks{~ Corresponding Authors}, Xianpei Han${}^{1, 2}$\footnotemark[1], Le Sun${}^{1,2}$\\
\textbf{Lingyong Yan}${}^{1,3}$, \textbf{Meng Liao}${}^{4}$, \textbf{Tong Xue}${}^{4}$, \textbf{Jin Xu}${}^{4}$\\
${}^{1}$Chinese Information Processing Laboratory ~ ${}^{2}$State Key Laboratory of Computer Science \\
Institute of Software, Chinese Academy of Sciences, Beijing, China\\
${}^{3}$University of Chinese Academy of Sciences, Beijing, China \\
${}^{4}$Data Quality Team, WeChat, Tencent Inc., China \\
{\tt \{boxi2020,hongyu,xianpei,sunle,lingyong2014\}@iscas.ac.cn} \\
{\tt \{maricoliao,xavierxue,jinxxu\}@tencent.com }}

\date{}

\begin{document}
\maketitle
\begin{abstract}
  Previous literatures show that pre-trained masked language models (MLMs) such as BERT can achieve competitive factual knowledge extraction performance on some datasets, indicating that MLMs can potentially be a reliable knowledge source. In this paper, we conduct a rigorous study to explore the underlying predicting mechanisms of MLMs over different extraction paradigms. By investigating the behaviors of MLMs, we find that previous decent performance mainly owes to the biased prompts which overfit dataset artifacts. Furthermore, incorporating illustrative cases and external contexts improve knowledge prediction mainly due to entity type guidance and golden answer leakage. Our findings shed light on the underlying predicting mechanisms of MLMs, and strongly question the previous conclusion that current MLMs can potentially serve as reliable factual knowledge bases\footnote{We openly release the source code and data at \url{https://github.com/c-box/LANKA}}.
\end{abstract}

\section{Introduction}

Recently, pre-trained language models~\citep{petersDeepContextualizedWord2018, devlinBERTPretrainingDeep2019,brownLanguageModelsAre2020}
have achieved promising performance on many NLP tasks. 
Apart from utilizing the universal representations from pre-trained models in downstream tasks, some literatures have shown the potential of pre-trained masked language models (e.g., BERT~\citep{devlinBERTPretrainingDeep2019} and  RoBERTa~\citep{liuRoBERTaRobustlyOptimized2019}) to be factual knowledge bases~\citep{petroniLanguageModelsKnowledge2019,bouraouiInducingRelationalKnowledge2019, jiangHowCanWe2020,shinAutoPromptElicitingKnowledge2020,jiangXFACTRMultilingualFactual2020,wangLanguageModelsAre2020,kassnerBERTkNNAddingKNN2020,kassnerArePretrainedLanguage2020a}. For example, to extract the birthplace of \textit{Steve Jobs}, we can query MLMs like BERT with ``\textit{Steve Jobs was born in [MASK]}'', where \textit{Steve Jobs} is the subject of the fact, \textit{``was born in''} is a prompt string for the relation ``\texttt{place-of-birth}'' and [MASK] is a placeholder for the object to predict.
Then MLMs are expected to predict the correct answer ``\textit{California}'' at the [MASK] position based on its internal knowledge. 
To help MLMs better extract knowledge, the query may also be enriched with external information like illustrative cases (\textit{e.g., (Obama, Hawaii)})~\citep{brownLanguageModelsAre2020} or external context (\textit{e.g., Jobs lives in California})~\citep{petroniHowContextAffects2020}. Some literatures have shown that such paradigms can achieve decent performance on some benchmarks like LAMA~\citep{petroniLanguageModelsKnowledge2019}.

\begin{figure}[tp]
  \centering
  \setlength{\belowcaptionskip}{-0.5cm}
  \includegraphics[width=0.45\textwidth]{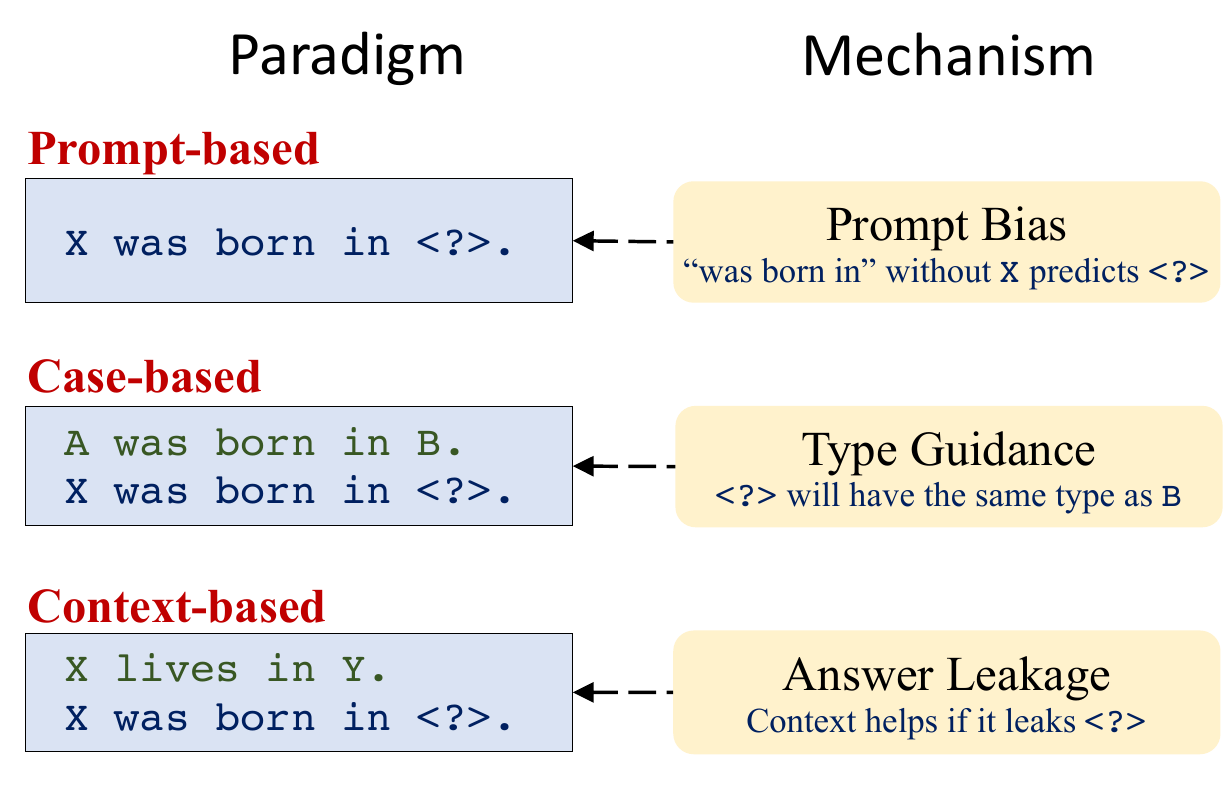}
  \caption{This paper explores three different kinds of factual knowledge extraction paradigms from MLMs, and reveal the underlying predicting mechanisms behind them.}
  \label{fig:main}
\end{figure}

Despite some reported success, currently there is no rigorous study looking deeply into the underlying mechanisms behind these achievements. Besides, it is also unclear whether such achievements depend on certain conditions (e.g., datasets, domains, relations). 
The absence of such kind of studies undermines our trust in the predictions of MLMs. We could neither determine whether the predictions are reliable nor explain why MLMs make a specific prediction, and therefore significantly limits MLMs' further applications and improvements. 

To this end, this paper conducts a thorough study on whether MLMs could be reliable factual knowledge bases.  Throughout our investigations, we analyze the behaviors of MLMs, figure out the critical factors for MLMs to achieve decent performance, and demonstrate how different kinds of external information influence MLMs' predictions. Specifically, we investigate factual knowledge extraction from MLMs\footnote{This paper shows the experimental results on BERT-large because previous work has shown that it can achieve the best performance on factual knowledge extraction among all MLMs. In the Appendix, we also report the experimental results on RoBERTa-large, which also reach the main conclusions reported in the paper.} over three representative factual knowledge extraction paradigms, as shown in Figure~\ref{fig:main}:
\begin{itemize}
  \item \textbf{Prompt-based retrieval}~\citep{petroniLanguageModelsKnowledge2019,jiangHowCanWe2020, shinAutoPromptElicitingKnowledge2020}, which queries MLM for object answer only given the subject and the corresponding relation prompt as input, e.g., \textit{``Jobs was born in [MASK].''}
  
  \item \textbf{Case-based analogy}~\citep{brownLanguageModelsAre2020,madottoLanguageModelsFewShot2020,gaoMakingPretrainedLanguage2020}, which enhances the prompt-based retrieval with several illustrative cases, e.g., \textit{``Obama was born in Hawaii. [SEP] Jobs was born in [MASK].''}
  
  \item \textbf{Context-based inference}~\citep{petroniHowContextAffects2020,bianBenchmarkingKnowledgeEnhancedCommonsense2021}, which augments the prompt-based retrieval with external relevant contexts, e.g., \textit{``Jobs lives in California. [SEP] Jobs was born in [MASK].''}
\end{itemize} 

Surprisingly, the main conclusions of this paper somewhat diverge from previous findings in published literatures, which are summarized in Figure~\ref{fig:main}. For prompt-based paradigm (\S~\ref{sec:prompt}), we find that the prediction distribution of MLMs is significantly prompt-biased. Specifically, we find that prompt-based retrieval generates similar predictions on totally different datasets. And predictions are spuriously correlated with the applied prompts, rather than the facts we want to extract. Therefore, previous decent performance mainly stems from the prompt over-fitting the dataset answer distribution, rather than MLMs’ knowledge extraction ability. Our findings strongly question the conclusions of previous literatures, and demonstrate that current MLMs can not serve as reliable knowledge bases when using prompt-based retrieval paradigm.

For case-based paradigm (\S~\ref{sec:example_based}), we find that the illustrative cases mainly provide a ``type guidance'' for MLMs. To show this, we propose a novel algorithm to induce the object type of each relation based on Wikidata\footnote{\url{www.wikidata.org}} taxonomy. According to the induced types, we find that the performance gain brought by illustrative cases mainly owes to the improvement on recognizing object type. By contrast, it cannot help MLMs select the correct answer from the entities with the same type: the rank of answer within its entity type is changed randomly after introducing illustrative cases. That is to say, under the case-based paradigm, although MLMs can effectively analogize between entities with the same type, they still cannot well identify the exact target object based on their internal knowledge and the provided illustrative cases.

For context-based paradigm (\S~\ref{sec:context_based}), we find that context can help the factual knowledge extraction mainly because it explicitly or implicitly leaks the correct answer. Specifically, the knowledge extraction performance improvement mainly happens when the introduced context contains the answer. Furthermore, when we mask the answer in the context, the performance still significantly improves as long as MLMs can correctly reconstruct the masked answer in the remaining context. In other words, in these instances, the context itself servers as a delegator of the masked answer, and therefore MLMs can still obtain sufficient implicit answer evidence even the answer doesn't explicitly appear.

All the above findings demonstrate that current MLMs are not reliable in factual knowledge extraction. Furthermore, this paper sheds some light on the underlying predicting mechanisms of MLMs, which can potentially benefit many future studies.

\section{Related Work}
\label{sec:background}

The great success of Pre-trained Language Models (PLMs) raises the question of whether PLMs can be directly used as reliable knowledge bases.~\citet{petroniLanguageModelsKnowledge2019} propose the LAMA benchmark, which probes knowledge in PLMs using prompt-based retrieval.~\citet{jiangXFACTRMultilingualFactual2020} build a multilingual knowledge probing benchmark based on LAMA. There are many studies focus on probing specific knowledge in PLMs, such as linguistic knowledge~\citep{linOpenSesameGetting2019,tenneyWhatYouLearn2019,liuLinguisticKnowledgeTransferability2019,htutAttentionHeadsBERT2019,hewittStructuralProbeFinding,goldbergAssessingBERTSyntactic2019,warstadtInvestigatingBERTKnowledge2019}, semantic knowledge~\citep{tenneyWhatYouLearn2019,wallaceNLPModelsKnow2019,ettingerWhatBERTNot2020} and world knowledge~\citep{feldmanCommonsenseKnowledgeMining2019,bouraouiInducingRelationalKnowledge2019,forbesNeuralLanguageRepresentations2019,zhouEvaluatingCommonsensePretrained2019,robertsHowMuchKnowledge2020a,linBirdsHaveFour2020,tamborrinoPretrainingAlmostAll2020a}. Recently, some studies doubt the reliability of PLMs as knowledge base by discovering the the spurious correlation to surface forms~\citep{mccoyRightWrongReasons2019,poernerEBERTEfficientYetEffectiveEntity2020,shwartzYouAreGrounded2020}, and their sensitivity to ``negation'' and ``mispriming''~\citep{kassnerNegatedMisprimedProbes2020b}.

Currently, there are three main paradigms for knowledge extraction from PLMs: prompt-based retrieval~\citep{schickExploitingClozeQuestions2020,liPrefixTuningOptimizingContinuous2021}, case-based analogy~\citep{schickFewShotTextGeneration2020,schickItNotJust2020a}, and context-based inference. For prompt-based retrieval, current studies focus on seeking better prompts by either mining from corpus~\citep{jiangHowCanWe2020} or learning using labeled data~\citep{shinAutoPromptElicitingKnowledge2020}. For case-based analogy, current studies mostly focus on whether good cases will lead to good few-shot abilities, and many tasks are tried~\citep{brownLanguageModelsAre2020, madottoLanguageModelsFewShot2020,gaoMakingPretrainedLanguage2020}. For context-based inference, current studies focus on enhancing the prediction by seeking more informative contexts, e.g., for knowledge extraction~\citep{petroniHowContextAffects2020} and CommonsenseQA~\citep{bianBenchmarkingKnowledgeEnhancedCommonsense2021}. However, there is no previous work which focuses on systematically study the underlying predicting mechanisms of MLMs on these paradigms.

\section{Prompt-based Retrieval}
\label{sec:prompt}

The prompt-based retrieval extracts factual knowledge by querying MLMs with (subject, prompt, [MASK]). For example, to extract the ``\texttt{place-of-birth}'' of \textit{Steve Jobs}, we could query BERT with ``\textit{Steve Jobs was born in [MASK].}'' and the predicted ``\textit{California}'' would be regarded as the answer. We consider three kinds of prompts: the manually prompts $T_{man}$ created by~\citet{petroniLanguageModelsKnowledge2019}, the mining-based prompts $T_{mine}$ by~\citet{jiangHowCanWe2020} and the automatically searched prompts $T_{auto}$ from ~\citet{shinAutoPromptElicitingKnowledge2020}.

\subsection{Overall Conclusion}

\begin{figure}[tp]
    \centering
    \begin{subfigure}[b]{0.48\textwidth}
        \centering
        \includegraphics[width=\textwidth]{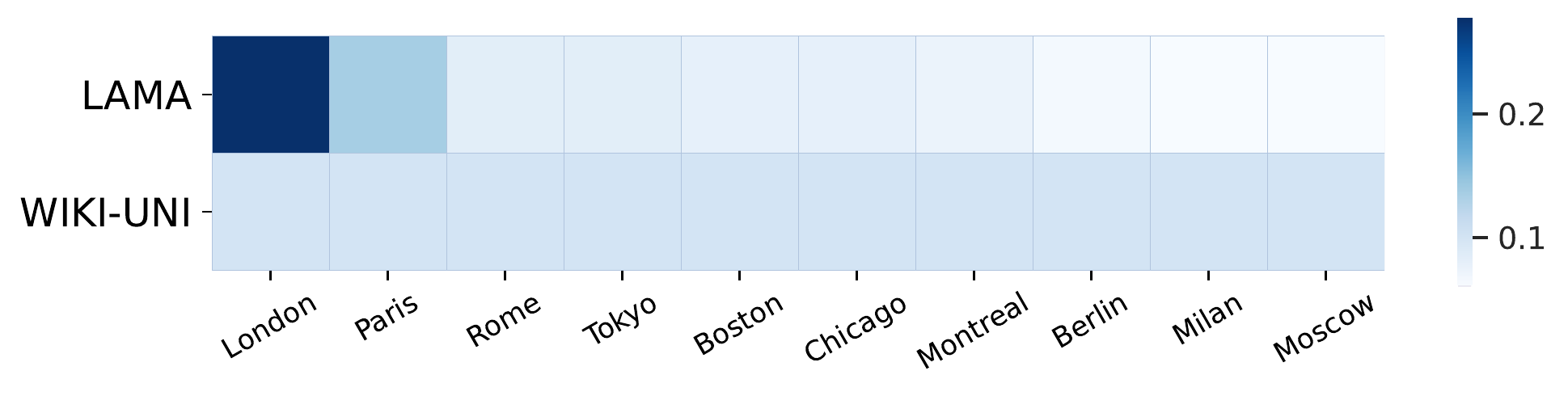}
        \caption{The true answer distributions are very different between LAMA and WIKI-UNI.}
        \label{subfig:obj_heat}
    \end{subfigure}

    \hfill

    \begin{subfigure}[b]{0.48\textwidth}
        \centering
        \includegraphics[width=\textwidth]{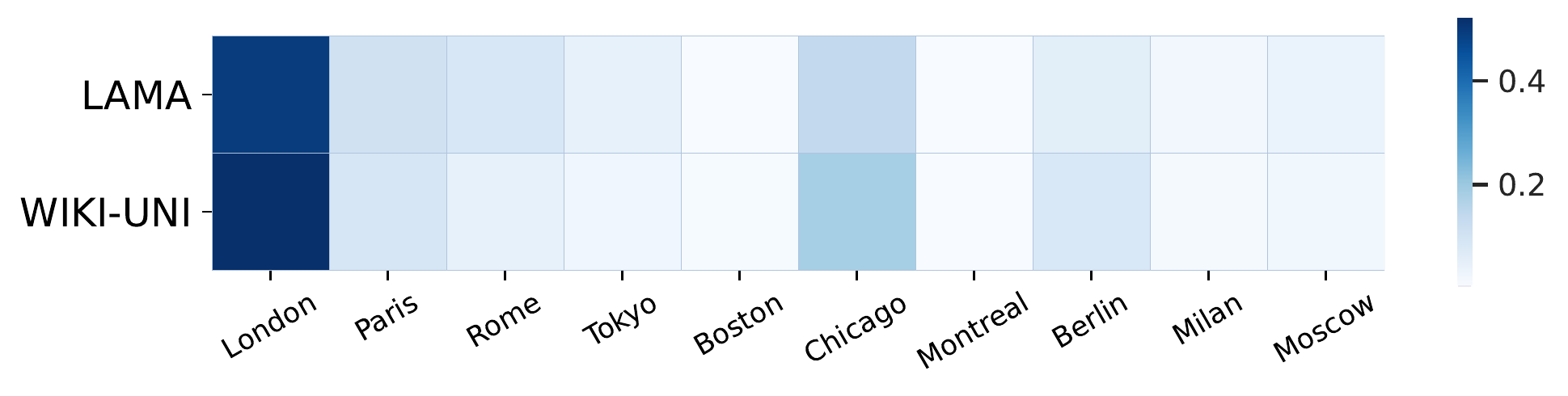}
        \caption{However, the prediction distribution made by MLMs on them are still very similar.}
        \label{subfig:predict_heat}
    \end{subfigure}
    \caption{An illustration example of the vastly different answer distributions but similar prediction distributions on LAMA and WIKI-UNI on ``\texttt{place-of-birth}'' relation.}
    \label{fig:dismap}
  \end{figure}

\textbf{Conclusion 1. } \emph{Prompt-based retrieval is prompt-biased. As a result, previous decent performance actually measures how well the applied prompts fit the dataset answer distribution, rather than the factual knowledge extraction ability from MLMs.}

Specifically, we conduct studies and find that 1) Prompt-based retrieval will generate similar responses given quite different datasets. To show this, we construct a new dataset from Wikidata -- WIKI-UNI, which have a totally different answer distribution from the widely-used LAMA\footnote{Since we focus on factual knowledge, we use the T-REx~\citep{elsaharTRExLargeScale} subset of the LAMA benchmark.} dataset~\citep{petroniLanguageModelsKnowledge2019}. However, we find that the prediction distributions on WIKI-UNI and LAMA are highly correlated, and this spurious correlation holds across different prompts. Such results reveal that there is just a weak correlation between the predictions of MLMs and the factual answer distribution of the dataset. 2) The prediction distribution is dominated by the prompt, i.e., the prediction distribution using only (prompt, [MASK]) is highly correlated to the prediction distribution using (subject, prompt, [MASK]). This indicates that it is the applied prompts, rather than the actual facts, determine the predictions of MLMs. 3) The performance of the prompt can be predicted by the divergence between the prompt-only distribution and the answer distribution of the dataset. All these findings reveal that previous decent performance in this field actually measures the degree of prompt-dataset fitness, rather than the universal factual knowledge extraction ability.

\begin{figure}[tp]
    \centering
    \setlength{\belowcaptionskip}{-0.3cm}
    \includegraphics[width=0.48\textwidth]{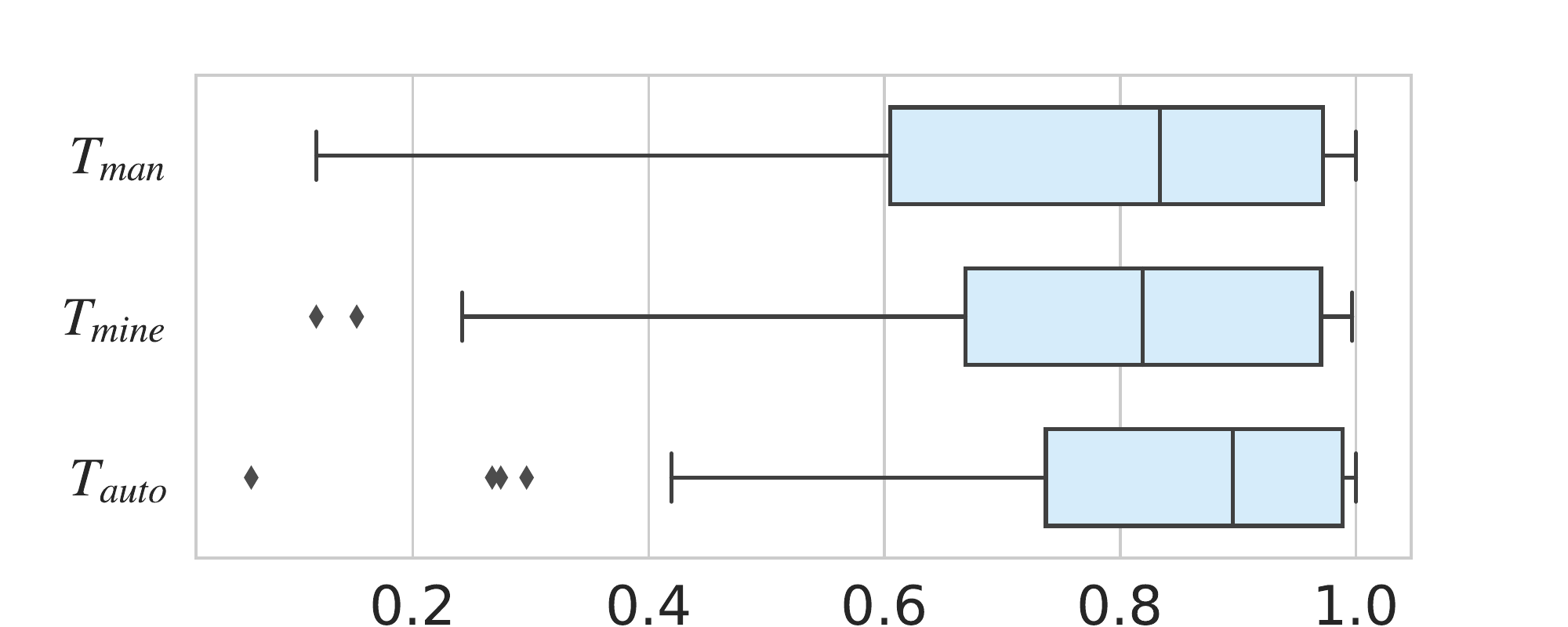}
    \caption{Correlations of the prediction distributions on LAMA and WIKI-UNI. Even these two datasets have totally different answer distributions, MLMs still make highly correlated predictions.}
    \label{fig:pre_corr}
\end{figure}

\subsection{Different Answers, Similar Predictions}
\label{sec:pb_f1}
\textbf{Finding 1.} \emph{Prompt-based retrieval will generate similar responses to quite different datasets.}

A reliable knowledge extractor should generate different responses to different knowledge queries. To verify whether MLMs meet this standard, we manually construct a new dataset -- WIKI-UNI, which has a comparable size but totally different answer distribution to LAMA, and then compare the prediction distributions on them. For a fair comparison, we follow the construction criteria of LAMA: we use the same 41 relations, filter out the queries whose objects are not in the MLMs’ vocabulary. Compared with LAMA, the major difference is that WIKI-UNI has a uniform answer distribution, i.e., for each relation, we keep the same number of instances for each object. Please refer to Appendix for more construction details. Figure~\ref{subfig:obj_heat} shows the answer distributions of LAMA and WIKI-UNI on relation “\texttt{place-of-birth}”. We can see that the answers in LAMA are highly concentrated on the head object entities, while the answers in WIKI-UNI follow a uniform distribution.

\begin{table}[tp]
    \centering
    \setlength{\belowcaptionskip}{-0.4cm}
    \resizebox{\columnwidth}{!}{
    \begin{tabular}{c|cccc|c}
        \toprule
        \textbf{Distribution}       & \textbf{Datasets} & \textbf{Top1} & \textbf{Top3} & \textbf{Top5} & \textbf{Precision} \\ \hline
        \multirow{2}{*}{Answer} & {\small LAMA}              & 22.04          & 39.37         & 48.03  &  -      \\ 
                                     & {\small WIKI-UNI}      & 1.68          & 5.03          & 7.78   &  -     \\ \hline
        \multirow{2}{*}{Prediction}  & {\small LAMA}              & 31.09         & 49.21         & 57.93  &  30.36     \\ 
                                     & {\small WIKI-UNI}      & 27.12         & 44.19         & 52.18  &  16.47     \\ \bottomrule
    \end{tabular}
    }
    \caption{Average percentage of instances being covered by top-k answers or predictions. For answer distribution, top-5 objects in LAMA cover 6.2 times of instances than that in WIKI-UNI, however, for prediction distribution, they are almost the same. As a result, the precision is significantly dropped in WIKI-UNI.}
    \label{tab:topk_cover}
\end{table}

Given LAMA and WIKI-UNI, we investigate the predicting behaviors of MLMs. Surprisingly, the prediction distributions on these two totally different datasets are highly correlated. Figure~\ref{subfig:predict_heat} shows an example. We can see that the prediction distribution on WIKI-UNI is very similar to that on LAMA. And these two distributions are both close to the answer distribution of LAMA but far away from the answer distribution of WIKI-UNI.

To investigate whether this spurious correlation is a common phenomenon, we analyze the Pearson correlation coefficient between prediction distributions on LAMA and WIKI-UNI across different relations and three kinds of prompts. The boxplot in Figure~\ref{fig:pre_corr} shows the very significant correlation between the prediction distributions on LAMA and WIKI-UNI: on all three kinds of prompts, the correlation coefficients exceed 0.8 in more than half of relations. These results demonstrate that prompt-based retrieval will lead to very similar prediction distributions even when test sets have vastly different answer distributions.

Furthermore, we find that the prediction distribution obviously doesn't correspond to the answer distribution of WIKI-UNI. From Table~\ref{tab:topk_cover}, we can see that on average, the top-5 answers of each relation in WIKI-UNI cover only 7.78\% instances. By contrast, the top-5 predictions of each relation in WIKI-UNI cover more than 52\% instances, which is close to the answer distribution and prediction distribution on LAMA. As a result, the performance on WIKI-UNI (mean P@1: 16.47) is significantly worse than that on LAMA (mean P@1: 30.36). In conclusion, the facts of a dataset cannot explain the predictions of MLMs, and therefore previous evaluations of the MLMs’ ability on factual knowledge extraction are unreliable.

\begin{figure}
    \centering
    \setlength{\belowcaptionskip}{-0.4cm}
    \includegraphics[width=0.48\textwidth]{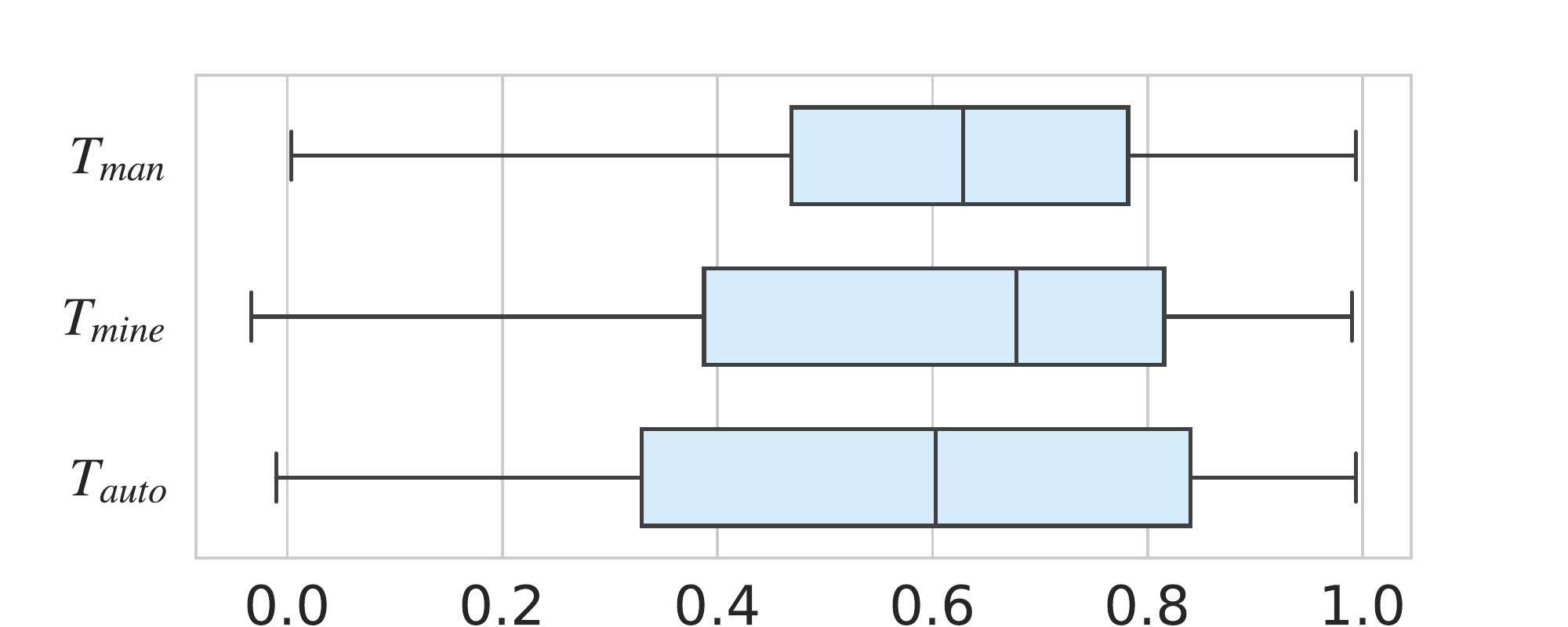}
    \caption{Correlations between the prompt-only distribution and prediction distribution on WIKI-UNI. MLMs make correlated predictions w. or w/o. subjects.}
    \label{fig:maskcorr}
\end{figure}

\subsection{Prompts Dominates Predictions}
\label{subsec:prompt_bias}

\textbf{Finding 2. } \emph{The prediction distribution is severely prompt-biased.}

To investigate the underlying factors of the predicting behavior of MLMs, we compare the prompt-only prediction distribution using only (prompt, [MASK]) and the full prediction distribution using (subject, prompt, [MASK]). To obtain the prompt-only distribution, we mask the subject and then use ([MASK], prompt, [MASK]) to query MLMs  (\textit{e.g., [MASK] was born in [MASK]}). Because there is no subject information in the input, MLMs can only depend on applied prompt's information to make the prediction at the second [MASK]. Therefore, we regard the probability distribution at the second [MASK] symbol as the prompt-only distribution.

After that, we analyze the correlations between the prompt-only distribution and the prediction distribution on WIKI-UNI dataset. Figure~\ref{fig:maskcorr} shows the boxplot. On all three kinds of prompts, correlation coefficients between the prompt-only distribution and the prediction distribution on WIKI-UNI exceed 0.6 in more than half of relations. This demonstrates that in these relations, the prompt-only distribution dominates the prediction distribution. Combining with the findings in Section~\ref{sec:pb_f1}, we can summarize that the prompt-based retrieval is mainly based on \emph{guided guessing}, i.e., the predictions are generated by sampling from the prompt-biased distribution guided by the moderate impact of subjects.

Note that among a minor part of relations, the correlations between the prompt-only distribution and the prediction distribution are relatively low. We find that the main reason is the type selectional preference provided by the subject entities, and Section~\ref{sec:example_based} will further discuss the impact of this type-guidance mechanism for MLMs.

\subsection{Better Prompts are Over-Fitting}
\label{subsec:prompt_overfit}
\textbf{Finding 3. } \emph{``Better'' prompts are the prompts fitting the answer distribution better, rather than the prompts with better retrieval ability.}

Some previous literatures attempt to find better prompts for factual knowledge extraction from MLMs. However, as we mentioned above, the prompt itself will lead to a biased prediction distribution. This raises our concern that whether the found better prompts are really with better knowledge extraction ability, or the better performance just come from the over-fitting between the prompt-only distribution and the answer distribution of the test set.

To answer this question, we evaluate the KL divergence between the prompt-only distribution and the answer distribution of LAMA on different kinds of prompts. The results are shown in Table~\ref{tab:pr_kl}. We find that the KL divergence is a strong indicator of the performance of a prompt, i.e., the smaller the KL divergence between the prompt-only distribution and the answer distribution of the test set is, the better performance the prompt achieve. Furthermore, Table~\ref{tab:overfit} shows several comparisons between different kinds of prompts and their performance on LAMA. We can easily observe that the better-performed prompts are actually over-fitting the dataset, rather than better capturing the underlying semantic of the relation. As a result, previous prompt searching studies are actually optimized on the spurious prompt-dataset compatibility, rather than the universal factual knowledge extraction ability.

\begin{table}[tp]
    \centering
    \begin{tabular}{ccc}
    \toprule
    \textbf{Prompt} & \textbf{Precision} & \textbf{KL divergence} \\ \hline
    $T_{man}$             &      30.36          &      12.27            \\
    $T_{mine}$            &      39.49          &      10.40            \\
    $T_{auto}$            &      40.36          &      10.27            \\ \bottomrule
    \end{tabular}
    \label{tab:kl}
    \caption{The smaller KL divergence between the prompt-only distribution and golden answer distribution of LAMA, the better performance of the prompt.}
    \label{tab:pr_kl}
\end{table}

\begin{table}[tp]
    \centering
    \setlength{\belowcaptionskip}{-0.4cm}
    \resizebox{\columnwidth}{!}{
    \begin{tabular}{llccc}
    \toprule
            \textbf{Relation}               & \textbf{Prompt}   & \textbf{Source}          & \textbf{Prec.}  &\textbf{KL.} \\ \hline
            \multirow{2}{*}{\texttt{citizenship}} & $x$ is $y$ citizen & $T_{man}$ & 0.00      &  24.67   \\
                                                        & $x$ returned to $y$ & $T_{mine}$ & 43.58  &   6.32      \\  \hline
            \multirow{2}{*}{\texttt{work location}}  & $x$ used to work in $y$ & $T_{man}$ & 11.01  &   19.07      \\ 
                                          & $x$ was born in $y$ & $T_{mine}$ & 40.25   &  2.21     \\                      \hline
            \multirow{2}{*}{\texttt{instance of}} & $x$ is a $y$ & $T_{man}$ & 30.15   &   22.98     \\ 
            & $x$ is a small $y$ & $T_{mine}$ & 52.60  &13.98        \\  \bottomrule   
    \end{tabular}
    }
    \caption{Examples of prompts that can achieve significant improvements on LAMA. We can see that the better performance actually stems from over-fitting: the better prompts are not prompts with a stronger semantic association to the relation.}
    \label{tab:overfit}
\end{table}

\section{Case-based Analogy}
\label{sec:example_based}
 
The case-based analogy enhances the prompt-based paradigm with several illustrative cases. For example, if we want to know the ``\texttt{place-of-birth}'' of \textit{Steve Jobs}, we would first sample cases such as (\textit{Obama}, \texttt{place-of-birth}, \textit{Hawaii}), and combine them with the original query. In this way, we will use ``\textit{Obama was born in Hawaii. [SEP] Steve Jobs was born in [MASK].}'' to query MLMs.

\subsection{Overall Conclusion}
\textbf{Conclusion 2.} \emph{Illustrative cases guide MLMs to better recognizing object type, rather than better predicting facts.}

To show this, we first design an effective algorithm to induce the type of an entity set based on Wikidata taxonomy, which can identify the object type of a relation. According to the induced types, we find that the benefits of illustrative cases mainly stem from the promotion of object type recognition. In other words, case-based analogy guides MLMs with better type prediction ability but contributes little to the entity prediction ability. In the following, we first illustrate our type inducing algorithm, and then explain how we reach the conclusion.

\subsection{Entity Set Type Induction}
\label{subsec:algorithm}

\begin{figure}
    \centering
    \includegraphics[width=0.49\textwidth]{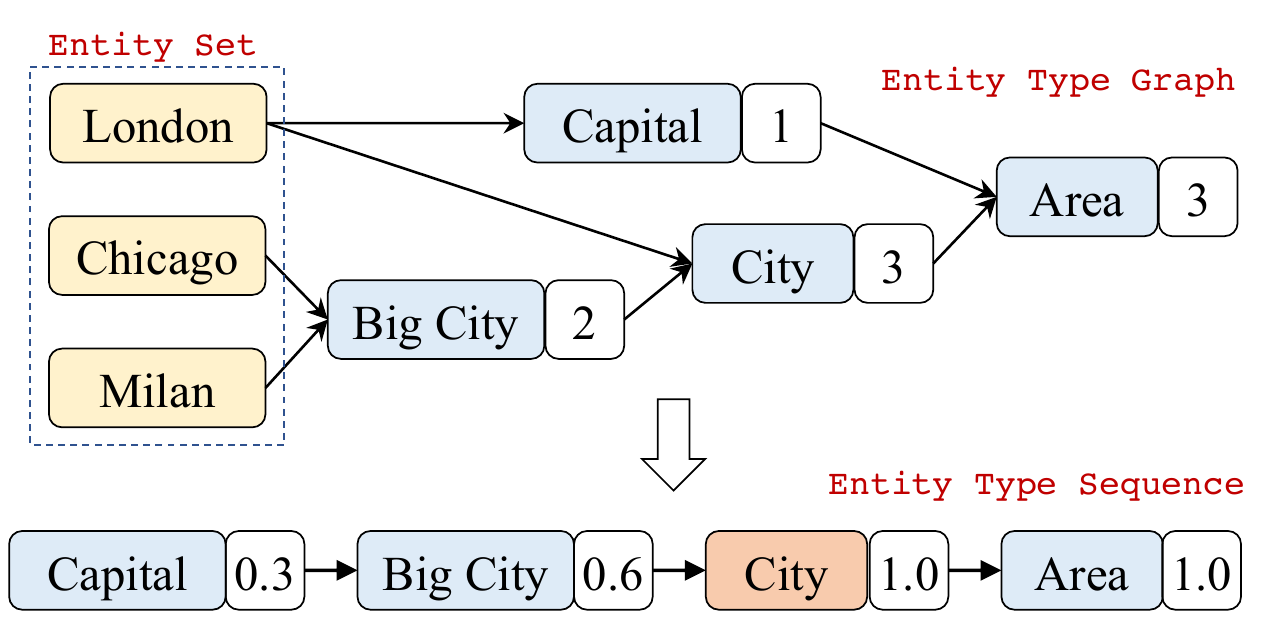}
    \caption{Illustration of our type induction algorithm. The numbers on the right of each type indicate how many entities does the type cover. The type of an entity set is the finest grained type in the type graph that can cover a sufficient number of the instances in the entity set, which is \emph{City} in the example.}
    \label{fig:entity_type}
  \end{figure}

  To induce the object type of a relation, we first collect all its objects in LAMA and form an entity set. Then we induce the type of an entity set by designing a simple but effective algorithm. The main intuition behind our algorithm is that a representative type should be  the finest grained type that can cover a sufficient number of the instances in the entity set. Figure \ref{fig:entity_type} shows an example of our algorithm. Given a set of entities in Wikidata, we first construct an entity type graph (ETG) by recursively introducing all ancestor entity types according to the \texttt{instance-of} and \texttt{subclass-of} relations. For the example in Figure \ref{fig:entity_type}, \textit{Chicago} is in the entity set and is an \texttt{instance-of} \textit{Big City}. \textit{Big City} is a \texttt{subclass-of} \textit{City}. As a result, \textit{Chicago}, \textit{Big City} and \textit{City} will all be introduced into ETG. Then we apply topological sorting \citep{COOK19852} to ETG to obtain a \emph{Fine-to-Coarse entity type sequence}. Finally, based on the sequence, we select the first type which covers more than 80\% of entities in the entity set (e.g., \textit{City} in Figure~\ref{fig:entity_type}). Table \ref{tab:ana_type} illustrates several induced types, and please refer to the Appendix for details.

\begin{table*}[tp]
  \centering
  \resizebox{0.9\textwidth}{!}{
  \begin{tabular}{l|l|ccc|c}
  \toprule
\textbf{Relation}   & \textbf{Induced Object Type}  & \textbf{\makecell[c]{Precision \\$\Delta$ }}& \textbf{\makecell[c]{Type \\ Prec. $\Delta$}} & \textbf{\makecell[c]{Wrong $\rightarrow$ Right \\ w/ Type Change}} & \textbf{\makecell[c]{Right $\rightarrow$ Wrong \\ w/o Type Change}}\\ \hline
  \texttt{country of citizenship}                           &         sovereign state      & 43.37          & 84.16          & 100.00                    & -                    \\
  \texttt{position held}                                    &         religious servant    & 36.88          & 80.26          & 91.15                     & 90.00                     \\
  \texttt{religion}                                         &          religion            & 33.20          & 34.88          & 100.00                    & -                      \\
  \texttt{work location}                                    &         city                 & 26.10          & 70.55          & 85.04                     & 100.00                    \\
  \texttt{instrument}                                       &         musical instrument   & 17.07          & 55.75          & 89.08                     & 75.00                     \\
  \texttt{country}                                          &      sovereign state         & 14.30          & 29.04          & 88.48                     & 87.93                     \\
  \texttt{employer}                                         &      business                & 12.01          & 99.22          & 100.00                    & -                    \\
  \texttt{continent}                                        &     continent                & 10.87          & 51.18          & 96.86                     & 88.24                     \\

  \bottomrule
  \end{tabular}
}
  \caption{Detailed analysis on relations where the mean precision increased more than 10\%.
  Precision $\Delta$ and Type Prec. $\Delta$ represents the precision changes on the answer and the type of the answer respectively. 
  ``w/ Type Change'' and ``w/o Type Change'' represents the type of prediction changed/unchanged before/after introducing illustrative cases. ``-'' indicate there is no queries whose predictions are mistakenly reversed.}
  \label{tab:ana_type}
\end{table*}

\begin{figure}
  \centering
  \setlength{\belowcaptionskip}{-0.4cm}
  \includegraphics[width=\columnwidth]{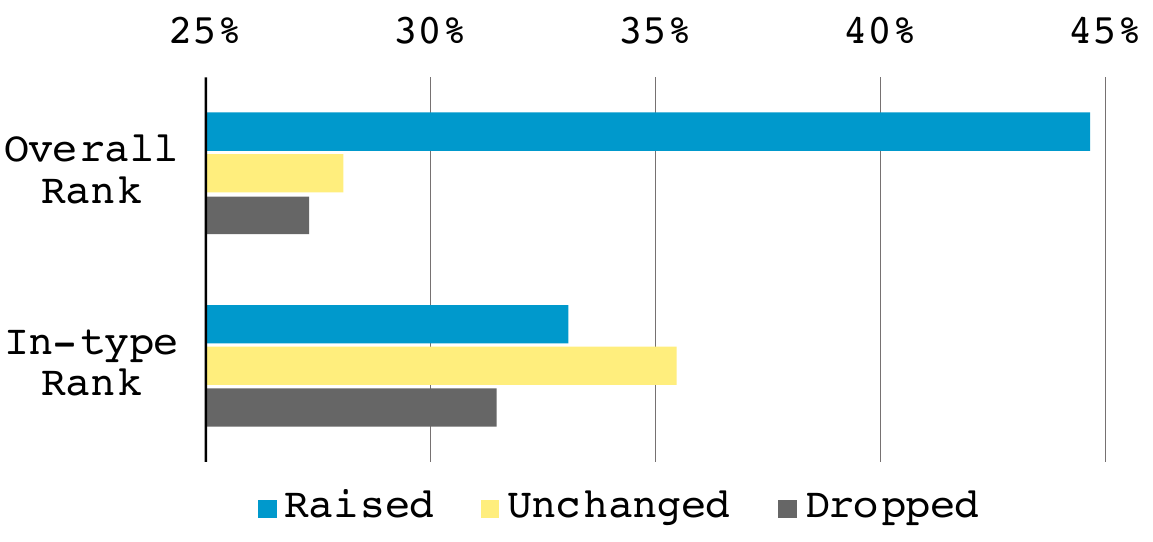}
  \caption{Percentages on the change of overall rank (among all candidates) and the in-type rank (among candidates with the same type) of golden answer. We can see that the illustrative cases mainly raise the overall rank but cannot raise the in-type rank, which means the performance improvements mainly come from better type recognition.}
  \label{fig:type_rank}
\end{figure}

\subsection{Cases Help Type Recognition}
\textbf{Finding 4. } \emph{Illustrative cases help MLMs to better recognize the type of objects, and therefore improve factual knowledge extraction.}

For case-based analogy, the first thing we want to know is whether illustrative cases can improve the knowledge extraction performance. To this end, for each (subject, relation) query in LAMA, we randomly sample 10 illustrative cases. To avoid answer leakage, we ensure the objects of these cases don’t contain the golden answer of the query. Then we use  (cases, subject, prompt, [MASK]) as the analogous query to MLMs.

Results show that case-based analogy can significantly improve performance. After introducing illustrative cases, the mean precision increases from 30.36\% to 36.23\%. Besides, we find that 11.81\% instances can benefit from the introduced cases and only 5.94\% instances are undermined. This shows that case-based analogy really helps the MLMs to make better predictions.

By analyzing the predicting behaviors, we observe that the main benefit of introducing illustrative cases comes from the better type recognition. To verify this observation, we investigate how the types of predictions changed after introducing the illustrative cases. Table \ref{tab:ana_type} shows the results on relations whose precision improvement is more than 10\% after introducing illustrative cases. From the table, it is very obvious that illustrative cases enhance the factual knowledge extraction by improving type prediction: 1) For queries whose predictions are correctly reversed (from wrong to right), the vast majority of them stems from the revised type prediction; 2) Even for queries whose predictions are mistakenly reversed (from right to wrong), the type of the majority of predictions still remains correct. In conclusion, introducing illustrative cases can significantly improve the knowledge extraction ability by recognizing the object type more accurately. That is, adding illustrative cases will provide more guidance for object type.

\subsection{Cases do not Help Entity Prediction}
\textbf{Finding 5. } \emph{Illustrative cases are of limited help for selecting the answer from entities of the same type.}

To show this, we introduce a new metric referred as \emph{in-type rank}, which is the rank of the correct answer within the entities of the same type for a query. By comparing the in-type rank in prompt-based and case-based paradigm, we can evaluate whether the illustrative cases can actually help better entity prediction apart from better type recognition.

Figure~\ref{fig:type_rank} shows the percentages on the change of overall rank (among all candidates) and the in-type rank (among candidates with the same type) of golden answer.
Unfortunately, we find that illustrative cases are of limited help for entity prediction: the change of in-type rank is nearly random. The percentages of queries with Raised/Unchanged/Dropped in-type rank are nearly the same: 33.05\% VS 35.47\% VS 31.47\%. 
Furthermore, we find that the MRR with the type only changed from 0.491 to 0.494, which shows little improvement after introducing illustrative cases.
These results show that the raises of overall rank of golden answer are not because of the better prediction inside the same type. 
In conclusion, illustrative cases cannot well guide the entity prediction, and they mainly benefit the factual knowledge extraction by providing guidance for object type recognition.

\section{Context-based Inference}
\label{sec:context_based}

The context-based inference augments the prompt-based paradigm with external contexts. For example, if we want to know the ``\texttt{place-of-birth}'' of \textit{Steve Jobs}, we can use the external context ``\textit{Jobs was from California.}'', and form a context-enriched query ``\textit{Jobs was from California. [SEP] Steve Jobs was born in [MASK].}'' to query MLMs.
Specifically, we use the same context retrieval method as \citet{petroniHowContextAffects2020}: for each instance, given the subject and relation as query, we use the first paragraph of DRQA's \citep{chenReadingWikipediaAnswer2017} retrieved document as external contexts.

\subsection{Overall Conclusion}
\textbf{Conclusion 3.} \emph{Additional context helps MLMs to predict the answer because they contain the answer, explicitly or implicitly.}

Several studies~\citep{petroniHowContextAffects2020, bianBenchmarkingKnowledgeEnhancedCommonsense2021} show that external context can help knowledge extraction from MLMs. To investigate the underlying mechanism, we evaluate which kinds of information in contexts contribute to the fact prediction, and find that the improvement mainly comes from the answer leakage in context. Furthermore, we find the answers can not only be leaked explicitly, but can also be leaked implicitly if the context provides sufficient information.

\subsection{Explicit Answer Leakage Helps}
\label{subsec:explicit}

\textbf{Finding 6. } \emph{Explicit answer leakage significantly improves the prediction performance.}
  \begin{table}[tp]
    \centering
    \setlength{\belowcaptionskip}{-0.5cm}
    \resizebox{\columnwidth}{!}{
    \begin{tabular}{cccc}
    \toprule
    \textbf{\makecell[c]{Answer \\ in context}} & \textbf{Prompt-based} & \textbf{Context-based} & $\Delta$ \\ \hline
    \makecell[c]{Present \\ (45.30\%) }
                                & 34.83       &     64.13        & +29.30       \\\hline
    \makecell[c]{Absent \\ (54.70 \%) }
                              & 25.37      &    23.26      & -2.11          \\
                                \bottomrule
    \end{tabular}
    }
    \caption{Comparison between prompt-based and context-based paradigms grouped by whether the answer presents or absents in the context. We can see that only contexts containing the answer can significantly improve the performance.}
    \label{tab:answer_pre}
    \end{table}

To show this, we split LAMA into two parts according to whether the additional context contains the answer. Table \ref{tab:answer_pre} shows the results on these two parts respectively. We can see that the improvements on these two parts diverge significantly. For context containing the answer, context-based inference significantly improves the factual knowledge extraction performance. However, there is even a little performance drop for those instances whose context does not contain the answer. This indicates that the improvement of factual knowledge extraction is mainly due to the explicit existence of the answer in the context.

\subsection{Implicit Answer Leakage Helps}
\textbf{Finding 7. } \emph{Implicit answer leakage can also significantly improve the prediction performance.}

\begin{table}
    \centering
    \resizebox{0.48\textwidth}{!}{
        \begin{tabular}{ccc}
            \toprule
            \textbf{Prompt-based} & \textbf{Context-based} & \textbf{Masked Context-based} \\ \hline
            30.36              & 41.44           & 35.66             \\ \bottomrule
            \end{tabular}
    }
    \caption{Overall performance when introducing different kinds of contexts. ``Masked Context-based'' indicates that we mask the golden answer in contexts, and there is still a significant performance improvement.}
    \label{tab:context_overall}
  \end{table}

  \begin{table}[tp]
    \centering
    \setlength{\belowcaptionskip}{-0.5cm}
    \resizebox{\columnwidth}{!}{
    \begin{tabular}{cccc}
    \toprule
    \textbf{\makecell[c]{Answer \\ Reconstructable}} & \textbf{Prompt-based} & \textbf{Context-based} & $\Delta$ \\ \hline
    \makecell[c]{Reconstructable \\ (60.23\%)}
                               & 39.58          &     60.82       & +21.24 \\ \hline
    \makecell[c]{Not-reconstructable \\ (39.77 \%) }
                              & 28.84       &    35.83     & +6.99              \\\bottomrule
    \end{tabular}
    }
    \caption{Comparison between prompt-based and context-based paradigms grouped by whether the masked answer in the context can be reconstructed from the remaining context. We can see that contexts can reconstruct the masked answer is more likely to improve the performance.}
    \label{tab:mask_context}
  \end{table}

As we mentioned above, explicit answer leakage significantly helps the answer prediction. The answer-leaked context may explicitly provide the answer or implicitly guide the prediction by providing answer-specific information. To understanding the underlying mechanism, we mask the answer in the context and verify whether it can still achieve the performance gain.

Table \ref{tab:context_overall} shows the results. We find that the performance gain is still very significant after masking the answer. This indicates that the contexts previously containing the answer are still very effective even the answer doesn't explicitly present.
To further investigate the reason behind, we split the masked version of answer-leaked instances into two groups by whether MLMs can or cannot correctly reconstruct the masked answer from the remaining context. The results are shown in Table \ref{tab:mask_context}. We can see that the performance gain significantly diverges in these two groups: the improvements mainly come from the instances whose answer in context can be reconstructed -- we refer to this as \emph{implicit answer leakage}. 
That is to say, for these instances, the context serves as a sufficient delegator of its answer, and therefore MLMs can obtain sufficient answer evidence even the answer does not explicitly appear. However, for contexts that cannot reconstruct the masked answer, the improvements are relatively minor. In conclusion, the real efficacy of context-based inference comes from the sufficient answer evidence provided by the context, either explicitly or implicitly.

\section{Conclusions and Discussions}
\label{sec:conclusions}

In this paper, we thoroughly study the underlying mechanisms of MLMs on three representative factual knowledge extraction paradigms. We find that the prompt-based retrieval is severely prompt-biased, illustrative cases enhance MLMs mainly via type guidance, and external contexts help knowledge prediction mostly because they contain the correct answer, explicitly or implicitly. These findings strongly question previous conclusions that current MLMs could serve as reliable factual knowledge bases.

The findings of this paper can benefit the community in many directions. By explaining the underlying predicting mechanisms of MLMs, we provide reliable explanations for many previous knowledge-intensive techniques. For example, our method can explain why and how incorporating external contexts will help knowledge extraction and CommonsenseQA~\citep{talmor2019commonsenseqa}. Our findings also reveal why PLM probing datasets may not be reliable and how the evaluation can be promoted by designing de-biased evaluation datasets.

This paper also sheds light on future research directions. For instance, knowing the main benefit of illustrative cases comes from type-guidance, we can enhance many type-centric prediction tasks such as NER~\citep{lample2016neural} and factoid QA~\citep{iyyer2014neural}. Moreover, based on the mechanism of incorporating external contexts, we can better evaluate, seek, and denoise external contexts for different tasks using MLMs. For example, we can assess and select appropriate facts for CommonsenseQA based on whether they can reconstruct the candidate answers.

This paper focuses on masked language models, which have been shown very effective and are widely used. We also want to investigate another representative category of language models --  the generative pre-trained models (e.g.,  GPT2/3~\citep{radfordLanguageModelsAre2019,brownLanguageModelsAre2020}), which have been shown to have quite different mechanisms and we leave it for future work due to page limitation.

\section*{Acknowledgments}
We sincerely thank all anonymous reviewers for their insightful comments and valuable suggestions.
This work is supported by the National Key Research and Development Program of China (No. 2020AAA0106400), the National Natural Science Foundation of China under Grants no. U1936207, and in part by the Youth Innovation Promotion Association CAS(2018141).


\bibliography{conf,addition}
\bibliographystyle{acl_natbib}

\clearpage

\appendix

\section{WIKI-UNI Construction Details}
\label{apx:dataset}

To construct WIKI-UNI, we first collect all the triples which belong to the same 41 relations with LAMA from Wikidata~\citep{10.1145/2629489}, then we randomly sample 50K triples with a single-token object for each relation. Similar to LAMA, we filter out the instances whose object is not in MLMs' vocabulary. For each relation, we group the instances based on different objects, and indicate $f_o$ as the frequency of each object. We denote the median of $f_o$ with $f_m$. For groups where $f_o>f_m$, we randomly sample $f_m$ instances, and delete the groups where $f_o<f_m$. Therefore, we acquire a dataset named WIKI-UNI with a uniform answer distribution. There are 70K facts in WIKI-UNI and 34K facts in LAMA. Since BERT and RoBERTa have a different vocabulary, so the datasets for their evaluation are slightly different. 

\section{Results on RoBERTa-large}
\label{apx:roberta}

Our conclusions are similar on BERT-large and RoBERTa-large, therefore, we report the results of BERT-large in the article and results of RoBERTa-large here. 

\subsection{Promp-based Retrieval}

\begin{figure}[h]
  \centering
  \includegraphics[width=0.48\textwidth]{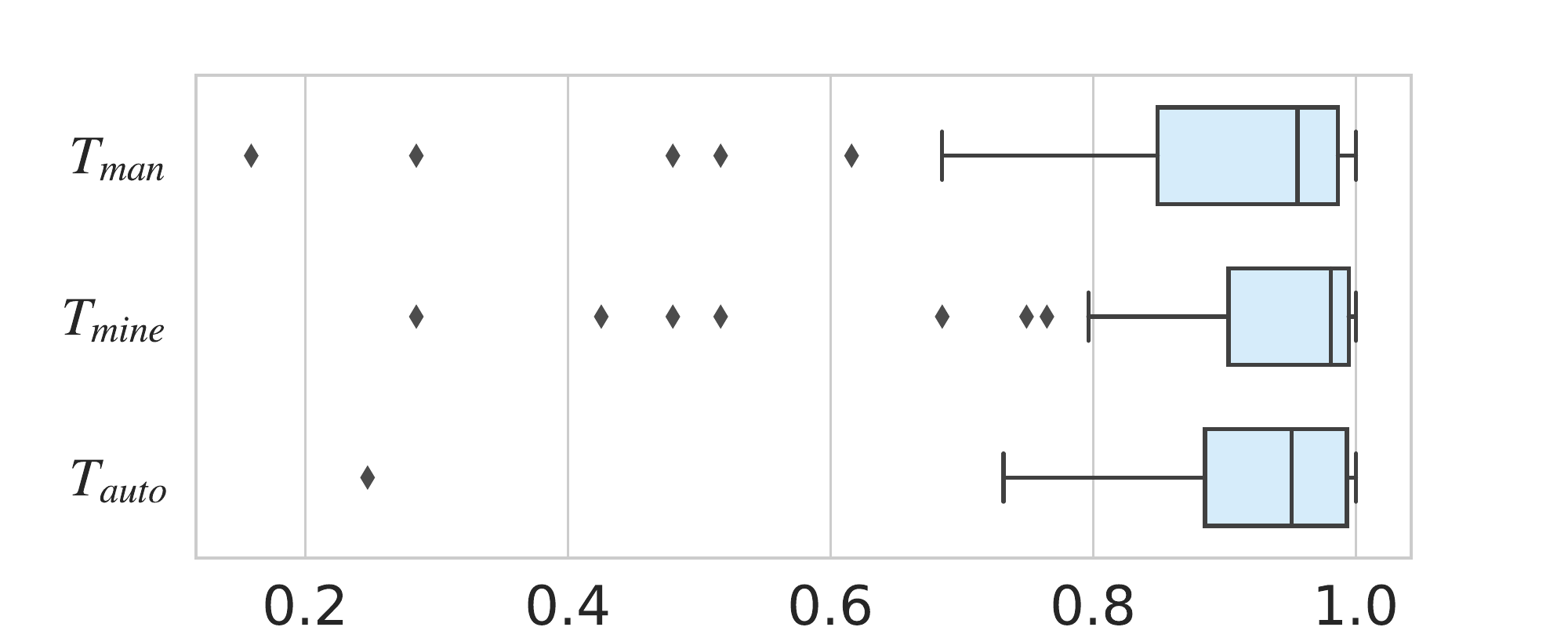}
  \caption{The correlations of the prediction distribution on LAMA and WIKI-UNI for RoBERTa-large.}
  \label{fig:roberta_pre_corr}
\end{figure}

Figure \ref{fig:roberta_pre_corr} shows the very significant correlation between the prediction distributions on LAMA and WIKI-UNI for RoBERTa-large: on all three kinds of prompts, the Pearson correlation coefficient between these two prediction distributions exceeds 0.9 in most relations.
Table \ref{tab:roberta_topk_cover} shows the percentage of instances that the topk object entities cover for RoBERTa-large. 

\begin{table}[h]
  \centering
  \resizebox{0.48\textwidth}{!}{
  \begin{tabular}{cc|ccc|c}
      \toprule
      \textbf{Distribution}       & \textbf{Datasets} & \textbf{Top1} & \textbf{Top3} & \textbf{Top5} & \textbf{Prec.} \\ \hline
      \multirow{2}{*}{Answer} & LAMA              & 23.93          & 42.02         & 50.08  &  -      \\ 
                                   & WIKI-UNI      & 1.84          & 5.53          & 8.61   &  -     \\ \hline
      \multirow{2}{*}{Prediction}  & LAMA              & 37.48         & 56.85         & 65.45  &  23.65     \\ 
                                   & WIKI-UNI      & 36.53         & 55.51         & 63.58  &  13.59     \\ \bottomrule
  \end{tabular}
  }
  \caption{The percentage of instances that the topk object entities cover for RoBERTa-large. The statistics is different from Table \ref{tab:topk_cover} because we filter LAMA with RoBERTa's vocabulary when evaluate RoBERTa-large.}
  \label{tab:roberta_topk_cover}
\end{table}

\subsection{Case-based Analogy}

\begin{figure}[h]
    \centering
    \includegraphics[width=\columnwidth]{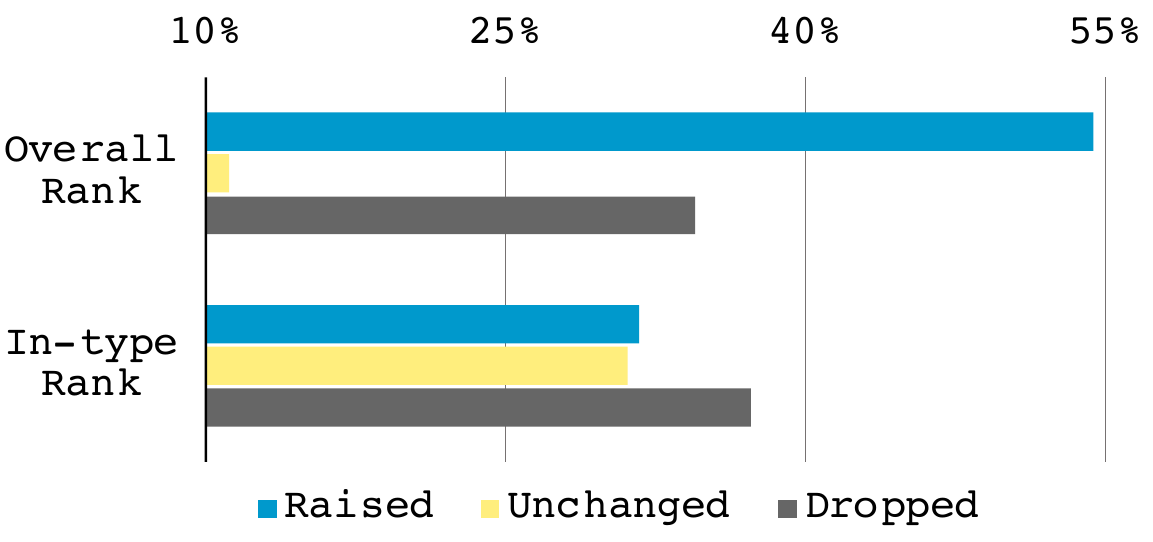}
    \caption{Percentages on the change of overall rank (among all candidates) and the in-type rank (among candidates with the same type) of golden answer of RoBERTa-large model.}
    \label{fig:type_rank_roberta}
  \end{figure}

Table \ref{tab:roberta_analogy} shows the performance improvement after introducing illustrative cases for RoBERTa-large model, we can see that the illustrative cases could also significantly increase the knowledge extraction performance for RoBERTa-large. 
Table \ref{fig:roberta_full_type} shows how the entity types of predictions changed after introducing the illustrative cases for RoBERTa-large model, the conclusion is similar with BERT-large. Figure \ref{fig:type_rank_roberta} shows the percentage on the change of overall rank and in-type rank for RoBERTa-large model.

And another finding is that BERT-large has a better type prediction ability than RoBERTa-large, even without illustrative cases. We calculate the overall type precision over prompt-based paradigm (the percentage of predictions that the type is correct).
And the type precision for BERT-large is 68\% and for RoBERTa-large is only 51\%, which partly explains why performance of RoBERTa-large is significantly worse than BERT-large on LAMA dataset. 

\begin{table}[h]
  \centering
  \begin{tabular}{cccc}
  \toprule
  \textbf{\makecell[c]{Enhanced with \\ Cases}} & \textbf{Prec.} & \textbf{Better} & \textbf{Worse} \\ \hline
  No                        &     23.65      &        -         &   -                          \\  
  Yes                       &     29.78      &     14.09       &     7.96                \\ \bottomrule 
  \end{tabular}
  \caption{Performance of the case-based analogy paradigm for RoBERTa-large}
  \label{tab:roberta_analogy}
\end{table}

\subsection{Context-based Inference}

Table \ref{tab:roberta_answer_pre} shows the comparison of contexts group by whether the contexts contain the answer for RoBERTa-large. We can see that for contexts containing the answer, context-based inference significantly improves the factual extraction performance. Meanwhile, there is a performance drop for those instances whose context does not contain the answer. Table \ref{tab:roberta_context_overall} shows the overall performance improvements when introducing different external contexts for RoBERTa-large. 
Table \ref{tab:roberta_reconstruct} shows the comparison of the masked contexts based on whether they can/cannot reconstruct the masked answer for RoBERTa-large. The improvements mainly comes from the instances whose answer in contexts can be reconstructed. 

\begin{table}[h]
  \centering
  \setlength{\belowcaptionskip}{-0.5cm}
  \resizebox{\columnwidth}{!}{
  \begin{tabular}{cccc}
  \toprule
  \textbf{\makecell[c]{Answer \\ in context}} & \textbf{Prompt-based} & \textbf{Context-based} & $\Delta$ \\ \hline
  \makecell[c]{Present \\ (46.04\%) }
                              & 27.95       &     52.05        & +24.10       \\\hline
  \makecell[c]{Absent \\ (53.96 \%) }
                            & 18.95      &    14.72      & -4.23          \\
                              \bottomrule
  \end{tabular}
  }
  \caption{Comparison of contexts grouped by whether the answer presents or absents for RoBERTa-large.}
\label{tab:roberta_answer_pre}
  \end{table}

\begin{table}[h]
  \centering
  \resizebox{0.48\textwidth}{!}{
      \begin{tabular}{ccc}
          \toprule
          \textbf{Without Contexts} & \textbf{Full Contexts} & \textbf{Masked Contexts} \\ \hline
          23.65              & 31.44           & 24.44             \\ \bottomrule
          \end{tabular}
  }
  \caption{The overall performance when introducing different contexts for RoBERTa-large.}
  \label{tab:roberta_context_overall}
\end{table}

\begin{table}[h]
  \centering
  \resizebox{\columnwidth}{!}{
  \begin{tabular}{cccc}
  \toprule
  \textbf{\makecell[c]{Answer \\ Reconstructable}} & \textbf{Prompt-based} & \textbf{Context-based} & $\Delta$ \\ \hline
  \makecell[c]{Reconstructable \\ (61.23\%)}
                             & 30.50          &     42.37       & +11.87  \\ \hline
  \makecell[c]{Not-reconstructable \\ (38.77 \%) }
                            & 22.19       &    22.15     & -0.04              \\\bottomrule
  \end{tabular}
  }
  \caption{Comparison of the masked contexts based on whether they can/cannot reconstruct the masked answer for RoBERTa-large.}
  \label{tab:roberta_reconstruct}
\end{table}

\section{Full Version of the Type Prediction Results}
  \label{apx:full}
  
  Table \ref{tab:full_type} shows the detailed analysis of all relations using case-based analogy paradigm for BERT-large and Table \ref{fig:roberta_full_type} is the results on RoBERTa-large. Because of the page limit, another finding we didn't mention in the article is that, apart from ``type guidance'', the illustrative cases could also provide a ``surface form guidance'' in a few relations (e.g., \texttt{part of}, \texttt{applies to jurisdiction}, \texttt{subclass of}). Specifically, the ``surface form'' indicate that the object entity name (\textit{e.g., Apple}) is a substring of the subject entity name (\textit{e.g., Apple Watch}). Such phenomenon is also mentioned in \citet{poernerEBERTEfficientYetEffectiveEntity2020}.

\clearpage

  \begin{table*}[h]
    \resizebox{0.98\textwidth}{!}{
    \begin{tabular}{l|l|ccc|c}
      \toprule
      \textbf{Relation}   & \textbf{Induced Object Type}  & \textbf{\makecell[c]{Precision \\$\Delta$ }}& \textbf{\makecell[c]{Type \\ Prec. $\Delta$}} & \textbf{\makecell[c]{Wrong $\rightarrow$ Right \\ w/ Type Change}} & \textbf{\makecell[c]{Right $\rightarrow$ Wrong \\ w/o Type Change}}\\ \hline
    named after                                      &         physical object      & 68.06          & 98.91          & 99.77                     & -                    \\
    country of citizenship                           &         sovereign state      & 43.37          & 84.16          & 100.00                    & -                    \\
    position held                                    &         religious servant    & 36.88          & 80.26          & 91.15                     & 90.00                     \\
    religion                                         &          religion            & 33.20          & 34.88          & 100.00                    & -                      \\
    work location                                    &         city                 & 26.10          & 70.55          & 85.04                     & 100.00                    \\
    instrument                                       &         musical instrument   & 17.07          & 55.75          & 89.08                     & 75.00                     \\
    country                                          &      sovereign state         & 14.30          & 29.04          & 88.48                     & 87.93                     \\
    employer                                         &      business                & 12.01          & 99.22          & 100.00                    & -                    \\
    continent                                        &     continent                & 10.87          & 51.18          & 96.86                     & 88.24                     \\
    languages spoken, written or signed              &     Indo-European languages  & 9.91           & -0.93          & 10.56                     & 81.54                     \\
    applies to jurisdiction                          &      state                   & 8.71           & -6.13          & 7.23                      & 63.64                     \\
    country of origin                                &     sovereign state          & 8.36           & 33.22          & 71.64                     & 98.28                     \\
    subclass of                                      &       object                 & 7.68           & 27.28          & 66.18                     & 87.10                     \\
    part of                                          &       object                 & 7.51           & 37.66          & 54.27                     & 97.87                     \\
    language of work or name                         &     Indo-European languages  & 6.05           & 10.95          & 77.23                     & 77.08                     \\
    location of formation                            &        city                  & 5.02           & 66.34          & 80.77                     & 100.00                    \\
    has part                                         &    abstract object           & 5.02           & 27.26          & 25.33                     & 100.00                    \\
    genre                                            &        series                & 4.62           & 17.61          & 95.45                     & -                      \\
    owned by                                         &       organization           & 2.62           & 11.50          & 9.57                      & 100.00                    \\
    instance of                                      &      concrete object         & 2.06           & 4.34           & 35.80                     & 96.77                     \\
    occupation                                       &        profession            & 1.35           & -0.53          & 0.00                      & 100.00                    \\
    place of death                                   &        city                  & 1.26           & 16.37          & 68.63                     & 100.00                    \\
    twinned administrative body                      &       city                   & 0.91           & 0.80           & 15.38                     & 75.00                     \\
    diplomatic relation                              &       sovereign state        & 0.80           & 1.11           & 10.00                     & 100.00                    \\
    native language                                  &    Indo-European languages   & 0.20           & 0.62           & 38.64                     & 92.86                     \\
    manufacturer                                     &       business               & -1.02          & 0.31           & 33.33                     & 61.29                     \\
    field of work                                    &       knowledge              & -1.15          & 0.00           & 26.09                     & 90.32                     \\
    developer                                        &       enterprise             & -1.52          & 1.52           & 4.17                      & 96.97                     \\
    location                                         &        community             & -1.57          & 4.59           & 3.03                      & 100.00                    \\
    capital                                          &        city                  & -2.00          & 0.14           & 4.55                      & 97.22                     \\
    position played on team / speciality             &         position             & -4.10          & 11.03          & -                    & 100.00                    \\
    headquarters location                            &         city                 & -4.24          & 0.62           & 0.00                      & 100.00                    \\
    official language                                &        Nostratic languages   & -5.28          & -1.14          & 5.45                      & 90.57                     \\
    original language of film or TV show             &        Nostratic languages   & -5.84          & -16.71         & 19.15                     & 43.30                     \\
    place of birth                                   &         city                 & -6.25          & 4.34           & 14.29                     & 100.00                    \\
    capital of                                       & political territorial entity & -6.84          & 0.42           & -                      & 100.00                    \\
    shares border with                               &          community           & -7.37          & 2.72           & 2.22                      & 97.35                     \\
    record label                                     &        record label          & -7.93          & -22.38         & -                    & 0.00                      \\
    original network                                 &      television station      & -10.56         & 0.45           & 11.36                     & 86.13                     \\
    located in the administrative territorial entity &      community               & -12.94         & 11.69          & 10.53                     & 99.25                     \\
    member of                                        &        organization          & -14.67         & 16.45          & 94.74                     & 98.08                     \\ \bottomrule            
    \end{tabular}
    }
    \caption{A detailed analysis of all relations using case-based analogy paradigm for BERT-large, which is corresponding to Table \ref{tab:ana_type} in the article. ``-'' indicates the number of queries whose predictions are reversed correctly or mistakenly is less than 3.}
    \label{tab:full_type}
    \end{table*}
  
    \begin{table*}[h]
      \resizebox{0.98\textwidth}{!}{
      \begin{tabular}{l|l|ccc|c}
        \toprule
        \textbf{Relation}   & \textbf{Induced Object Type}  & \textbf{\makecell[c]{Precision \\$\Delta$ }}& \textbf{\makecell[c]{Type \\ Prec. $\Delta$}} & \textbf{\makecell[c]{Wrong $\rightarrow$ Right \\ w/ Type Change}} & \textbf{\makecell[c]{Right $\rightarrow$ Wrong \\ w/o Type Change}}\\ \hline
        religion                                         &  religion  & 56.92  & 66.36  & 100.00           & -            \\
        position held                                    &  religious servant  & 41.86  & 47.42  & 99.03            & -             \\
        country of citizenship                           &  sovereign state & 37.16  & 74.11  & 100.00           & -           \\
        member of                                        &  organization  & 31.03  & 77.83  & 100.00           & -           \\
        continent                                        &  continent  & 29.51  & 87.80  & 100.00           & 100.00           \\
        instrument                                       &  musical instrument  & 28.26  & 6.04   & 94.04            & 0.00             \\
        country of origin                                &  sovereign state  & 28.18  & 94.92  & 99.61            & 100.00           \\
        country                                          &  sovereign state  & 26.64  & 69.84  & 95.22            & 96.55            \\
        part of                                          &  object  & 24.57  & 90.22  & 96.98            & 100.00           \\
        place of death                                   &  city  & 22.88  & 95.35  & 98.95            & 100.00           \\
        instance of                                      &  concrete object  & 14.97  & 20.53  & 34.30            & 97.50            \\
        location of formation                            &  city  & 14.12  & 99.88  & 100.00           & -           \\
        subclass of                                      &  object  & 12.07  & 26.25  & 63.31            & 90.00            \\
        capital                                          &  city  & 10.62  & 36.31  & 92.19            & 85.71            \\
        named after                                      &  physical object  & 10.25  & 85.05  & 100.00           & 100.00           \\
        language of work or name                         &  Indo-European languages  & 9.10   & 26.72  & 89.12            & 72.17            \\
        has part                                         &  abstract object  & 8.79   & 67.99  & 77.65            & -           \\
        work location                                    &  city  & 8.09   & 12.43  & 96.95            & 6.45             \\
        languages spoken, written or signed              &  Indo-European languages  & 5.09   & 17.75  & 54.20            & 86.90            \\
        employer                                         &  business  & 3.97   & 10.31  & 19.05            & 100.00           \\
        position played on team / speciality             &  position  & 3.26   & 56.51  & 71.43            & 75.00            \\
        native language                                  &  Indo-European languages  & 1.09   & 1.63   & 28.21            & 93.10            \\
        genre                                            &  series  & 1.05   & 0.23   & 75.00            & 66.67            \\
        record label                                     &  record label  & 0.00   & -7.55  & -           & -           \\
        place of birth                                   &  city  & -0.13  & 41.02  & 66.67            & 100.00           \\
        twinned administrative body                      &  city  & -0.45  & 1.04   & 0.00             & 100.00           \\
        headquarters location                            &  city  & -1.00  & 0.00   & 0.00             & 100.00           \\
        diplomatic relation                              &  sovereign state  & -1.16  & 1.05   & 25.00            & 100.00           \\
        owned by                                         &  organization  & -1.45  & 43.78  & 64.62            & 94.59            \\
        field of work                                    &  knowledge  & -2.10  & 0.69   & 10.53            & 96.77            \\
        occupation                                       &  profession  & -2.43  & 0.00   & 0.00             & 100.00           \\
        official language                                &  Nostratic languages  & -3.11  & 3.88   & 18.37            & 97.40            \\
        located in the administrative territorial entity &  community  & -3.35  & 45.81  & 75.93            & 97.50            \\
        original language of film or TV show             &  Nostratic languages  & -5.29  & -21.30 & 15.38            & 34.29            \\
        shares border with                               &  community  & -9.82  & 0.16   & 0.00             & 98.86            \\
        location                                         &  community  & -11.49 & 27.15  & 41.43            & 100.00           \\
        developer                                        &  enterprise  & -12.25 & 6.80   & 37.50            & 79.41            \\
        original network                                 &  television station  & -16.46 & -15.84 & 14.29            & 72.49            \\
        applies to jurisdiction                          &  state  & -18.38 & 2.11   & 35.71            & 98.00            \\
        capital of                                       &  political territorial entity  & -39.44 & 7.22   & -             & 100.00           \\
        manufacturer                                     &  business  & -49.63 & 6.79   & 44.44            & 93.82                                
        \\ \bottomrule            
      \end{tabular}
      }
      \caption{A detailed analysis of all relations using case-based analogy paradigm for RoBERTa-large, which is corresponding to Table \ref{tab:ana_type} in the article. ``-'' indicates the number of queries whose predictions are reversed correctly or mistakenly is less than 3.}
      \label{fig:roberta_full_type}
      \end{table*}

\end{document}